\documentclass[conference]{IEEEtran}
\IEEEoverridecommandlockouts
\usepackage{cite}
\usepackage{amsmath,amssymb,amsfonts}
\usepackage{algorithmic}
\usepackage{graphicx}
\usepackage{textcomp}
\usepackage{xcolor}
\usepackage{multirow}
\usepackage[caption=false]{subfig}
\usepackage{bm}
\def\BibTeX{{\rm B\kern-.05em{\sc i\kern-.025em b}\kern-.08em
    T\kern-.1667em\lower.7ex\hbox{E}\kern-.125emX}}

\newcommand{\IS}{IS$\uparrow$}
\newcommand{\FID}{FID$\downarrow$}
\newcommand{\etal}{\textit{et al.}}
\newcommand{\ie}{\textit{i.e.}}
\newcommand{\eg}{\textit{e.g.}}
\newcommand{\bmx}{{\bm x}}
\newcommand{\bmz}{{\bm z}}

\newcommand{\sentemb}{{\bm s}}

\newcommand{\E}{\mathop{\mathbb{E}}}

\newcommand{\image}{I}
\newcommand{\sentence}{S}
\newcommand{\m}{\mathbf{m}}

\newcommand{\si}{\textsc{SI}}

\newcommand{\genfromtext}[1]{$\phi(\text{``\textit{#1}"})$}

\newcommand{\figteaser}{
\newcommand{\h}{20.7mm}
\newcommand{\hh}{2.5mm}
\newcommand{\vv}{\vspace*{-0.35mm}}
\begin{figure}[t]
\centering
\begin{minipage}[c]{0.972\linewidth}
\hspace{3.0mm}%
\parbox[c]{\h}{\centering \footnotesize{There is a red bird with black beady eyes and dark edged feather sitting on a branch.}}%
\parbox[r]{\h}{\centering \footnotesize{This is a small yellow bird with a grey head and\\a small pointy beak.}}%
\parbox[l]{\h}{\centering \footnotesize{A flower with thin long pink petals and central cluster of orange stamen.}}%
\parbox[l]{\h}{\centering \footnotesize{The petals of the flower are color yellow with red stripes.}}\vspace{2mm}\\
\makebox[\hh]{\rotatebox[origin=l]{90}{\hspace{-1.18mm}\makebox[\h][c]{\footnotesize{Ours}}}}\hspace{0.5mm}%
\includegraphics[height=\h]{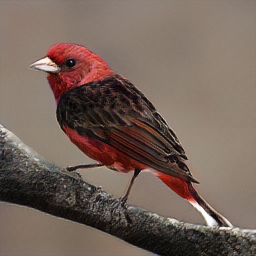}%
\includegraphics[height=\h]{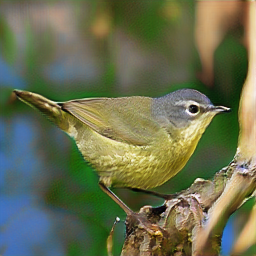}%
\includegraphics[height=\h]{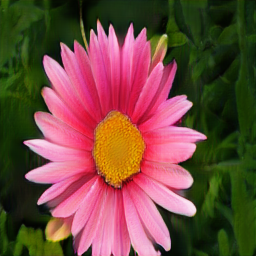}%
\includegraphics[height=\h]{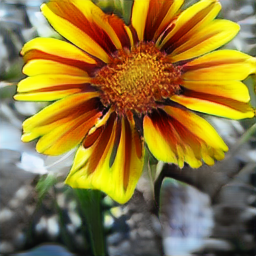}\vv\\
\makebox[\hh]{\rotatebox[origin=l]{90}{\hspace{-1.18mm}\makebox[\h][c]{\footnotesize{Ground Truth}}}}\hspace{0.5mm}%
\includegraphics[height=\h]{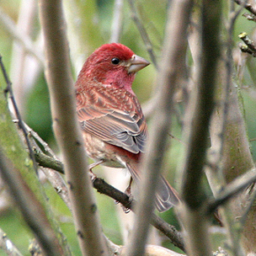}%
\includegraphics[height=\h]{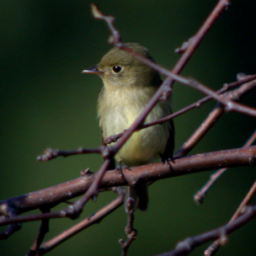}%
\includegraphics[height=\h]{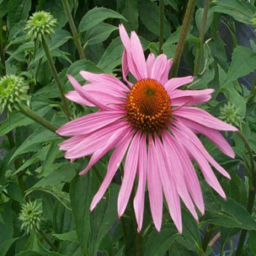}%
\includegraphics[height=\h]{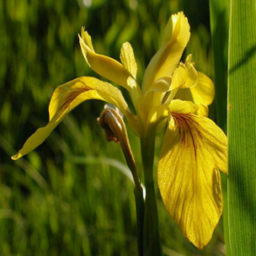}\vv\\
\end{minipage}
\caption{\label{fig:teaser}%
Images generated by our method.
\vspace*{-1mm}
}
\end{figure}
}

\newcommand{\figbirds}{
\renewcommand{\h}{27.4mm}
\renewcommand{\hh}{2.5mm}
\renewcommand{\vv}{\vspace*{-0.35mm}}
\begin{figure*}[!t]
\centering
\begin{minipage}[c]{167.4mm}
\hspace{3.0mm}%
\parbox[c]{\h}{\centering \footnotesize{An entirely black bird with small yellow eyes and a short straight bill.}}%
\parbox[r]{\h}{\centering \footnotesize{A blue bird with black legs and a short pointed beak.}}%
\parbox[l]{\h}{\centering \footnotesize{This white colored bird has bright orange feet and a hint of orange in its beak.}}%
\parbox[l]{\h}{\centering \footnotesize{This is a small, yellow bird with black on the crown, nape, and wingbars.}}%
\parbox[l]{\h}{\centering \footnotesize{This is a brown bird with a white breast and a large beak.

}}%
\parbox[l]{\h}{\centering \footnotesize{This colorful bird has a red crown adn throat with a black eye ring, and a white and pink belly.
}}\vspace{2mm}\\
\makebox[\hh]{\rotatebox[origin=l]{90}{\hspace{-1.18mm}\makebox[\h][c]{\normalsize{StackGAN++}}}}\hspace{0.5mm}%
\includegraphics[height=\h]{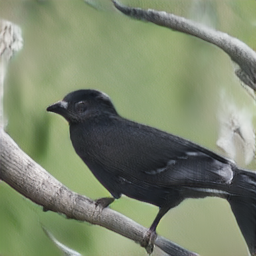}%
\includegraphics[height=\h]{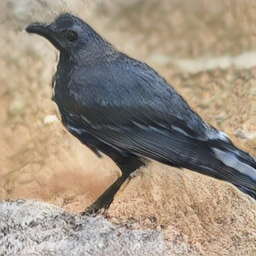}%
\includegraphics[height=\h]{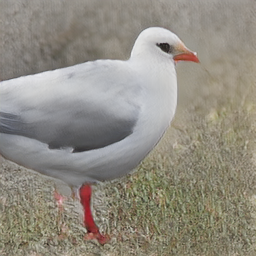}%
\includegraphics[height=\h]{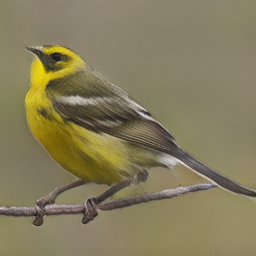}%
\includegraphics[height=\h]{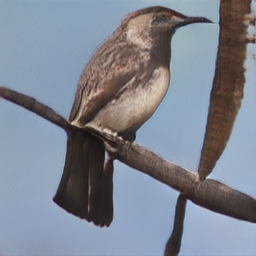}%
\includegraphics[height=\h]{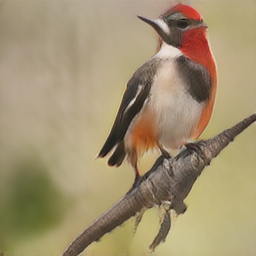}\vv\\
\makebox[\hh]{\rotatebox[origin=l]{90}{\hspace{-1.18mm}\makebox[\h][c]{\normalsize{HDGAN}}}}\hspace{0.5mm}%
\includegraphics[height=\h]{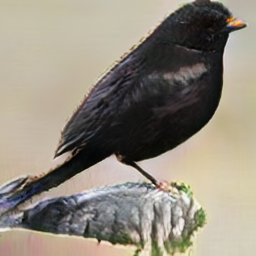}%
\includegraphics[height=\h]{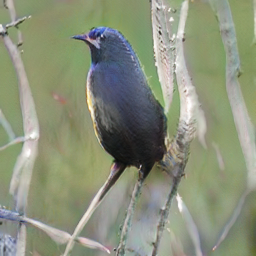}%
\includegraphics[height=\h]{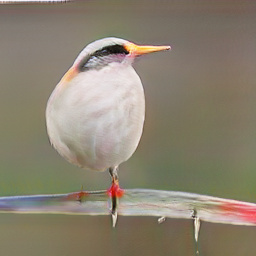}%
\includegraphics[height=\h]{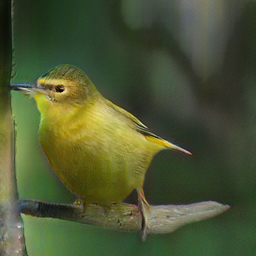}%
\includegraphics[height=\h]{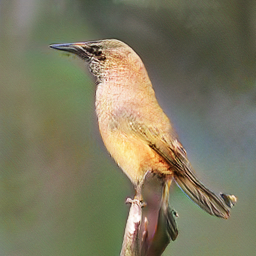}%
\includegraphics[height=\h]{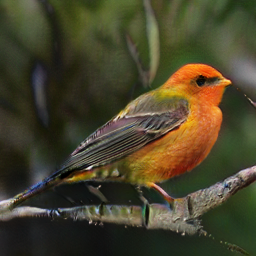}\vv\\
\makebox[\hh]{\rotatebox[origin=l]{90}{\hspace{-1.18mm}\makebox[\h][c]{\normalsize{Ours}}}}\hspace{0.5mm}%
\includegraphics[height=\h]{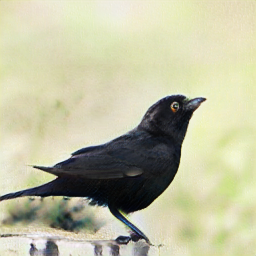}%
\includegraphics[height=\h]{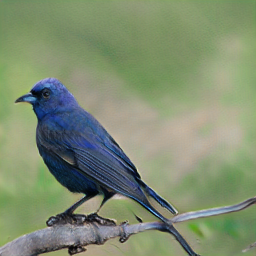}%
\includegraphics[height=\h]{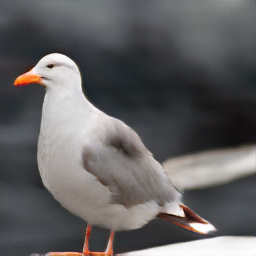}%
\includegraphics[height=\h]{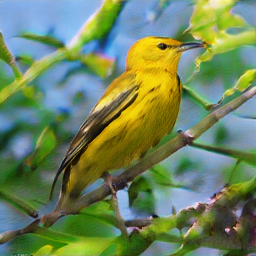}%
\includegraphics[height=\h]{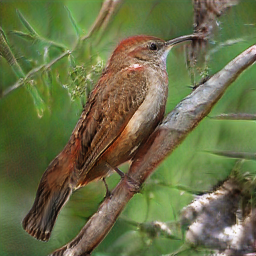}%
\includegraphics[height=\h]{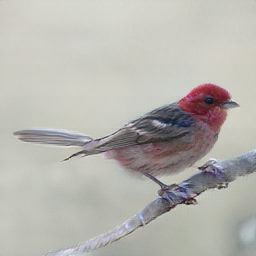}\vv\\
\end{minipage}
\caption{\label{fig:compare_birds}%
Qualitative results in the CUB Dataset.
\vspace*{-1mm}
}
\end{figure*}
}

\newcommand{\figflowers}{
\renewcommand{\h}{27.4mm}
\renewcommand{\hh}{2.5mm}
\renewcommand{\vv}{\vspace*{-0.35mm}}
\begin{figure*}[!t]
\label{fig:flowers}
\centering
\begin{minipage}[c]{167.4mm}
\hspace{3.0mm}%
\parbox[c]{\h}{\centering \footnotesize{This flower has petals that are yellow and folded together.}}%
\parbox[r]{\h}{\centering \footnotesize{This flower features elongated pointed orange petals emanating out of the main bulb.}}%
\parbox[l]{\h}{\centering \footnotesize{The flower has petals that are purple and white with purple filaments.}}%
\parbox[l]{\h}{\centering \footnotesize{This flower is pink in color, and has petals that are striped.}}%
\parbox[l]{\h}{\centering \footnotesize{This flower has wide and very smooth white petals with yellow central accents.}}%
\parbox[l]{\h}{\centering \footnotesize{The petals of the flower are in various colors such as red, yellow, and purple.}}\vspace{2mm}\\
\makebox[\hh]{\rotatebox[origin=l]{90}{\hspace{-1.18mm}\makebox[\h][c]{\normalsize{HDGAN}}}}\hspace{0.5mm}%
\includegraphics[height=\h]{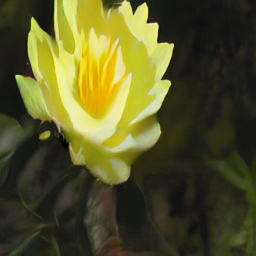}%
\includegraphics[height=\h]{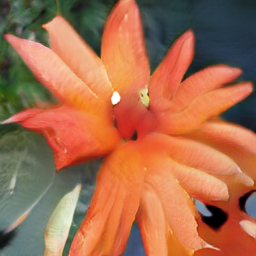}%
\includegraphics[height=\h]{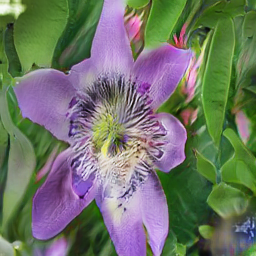}%
\includegraphics[height=\h]{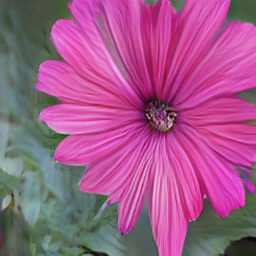}%
\includegraphics[height=\h]{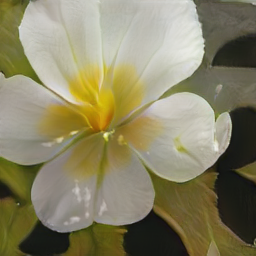}%
\includegraphics[height=\h]{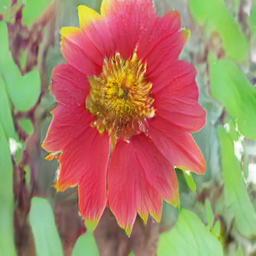}\vv\\
\makebox[\hh]{\rotatebox[origin=l]{90}{\hspace{-1.18mm}\makebox[\h][c]{\normalsize{Ours}}}}\hspace{0.5mm}%
\includegraphics[height=\h]{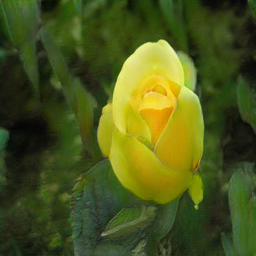}%
\includegraphics[height=\h]{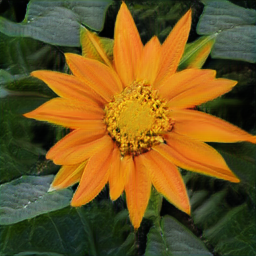}%
\includegraphics[height=\h]{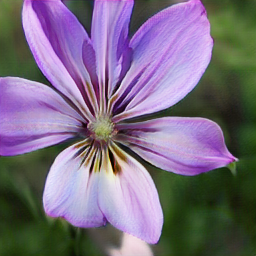}%
\includegraphics[height=\h]{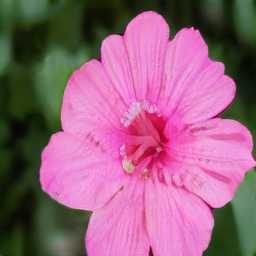}%
\includegraphics[height=\h]{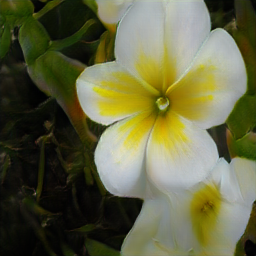}%
\includegraphics[height=\h]{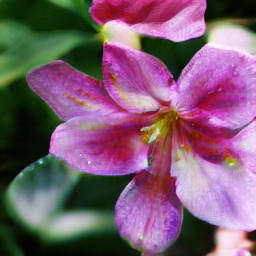}\vv\\
\end{minipage}
\caption{\label{fig:compare_flowers}%
Qualitative results in the Oxford-102 Dataset.
\vspace*{-1mm}
}
\end{figure*}
}

\newcommand{\figarithmetic}{
\renewcommand{\h}{22mm}
\renewcommand{\hh}{2.5mm}
\renewcommand{\vv}{\vspace*{-0.35mm}}
\begin{figure}[t]
\label{fig:vec_arithmetics}
\centering
\begin{minipage}[c]{68mm}
\parbox[c]{\h}{\centering \scriptsize{
$\phi (\mbox{``\textit{This is a}}$ \\ $\mbox{\textit{red bird}"})$
}}%
\parbox[r]{\h}{\centering \scriptsize{$-$ \genfromtext{It is red}}}%
\parbox[l]{\h}{\centering \scriptsize{$+$ \genfromtext{It is blue}}}%
\makebox[\hh][t]{}\hspace{0.5mm}%
\hspace{1.0mm}%
\includegraphics[height=\h]{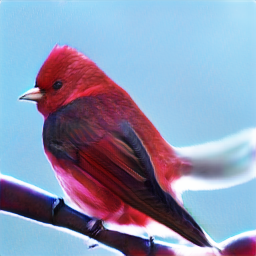}%
\hspace{1.0mm}%
\includegraphics[height=\h]{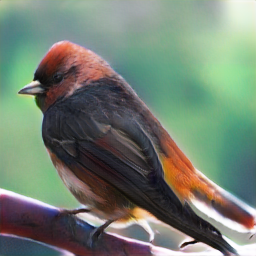}%
\hspace{1.0mm}%
\includegraphics[height=\h]{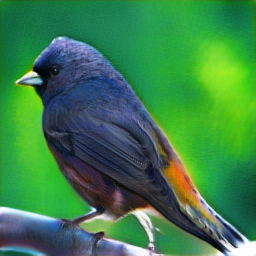}%
\hspace{1.0mm}%
\\
\vspace{1mm}
\parbox[t]{\h}{\centering \scriptsize{
     $\phi (\mbox{``\textit{Blue bird has}}$ \\ $\mbox{\textit{long beak}"})$
}}%
\parbox[c]{\h}{\centering \scriptsize{$-$ \genfromtext{Long beak}}}%
\parbox[l]{\h}{\centering \scriptsize{$+$ \genfromtext{Small beak}}}%
\\
\includegraphics[height=\h]{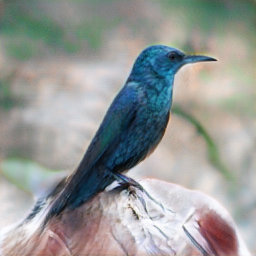}%
\hspace{1.0mm}%
\includegraphics[height=\h]{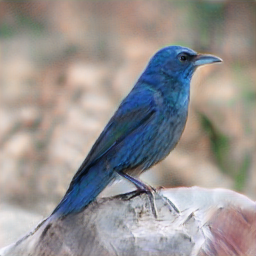}%
\hspace{1.0mm}%
\includegraphics[height=\h]{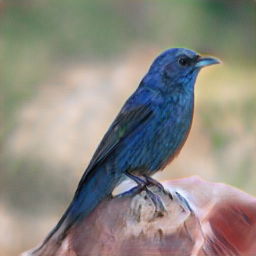}%
\\
\vspace{1mm}
\parbox[t]{\h}{\centering \scriptsize{
$\phi (\mbox{``\textit{White bird}}$ \\ $\mbox{\textit{long beak}"})$
}}%
\parbox[c]{\h}{\centering \scriptsize{$-$ \genfromtext{Red crown}}}%
\parbox[l]{\h}{\centering \scriptsize{$+$ \genfromtext{Black crown}}}%
\\
\includegraphics[height=\h]{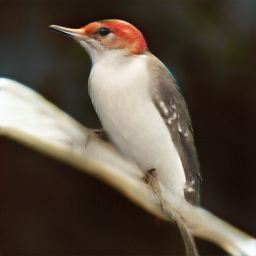}%
\hspace{1.0mm}%
\includegraphics[height=\h]{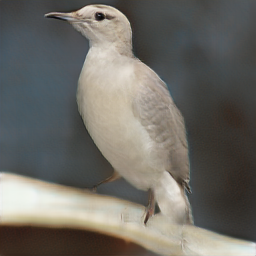}%
\hspace{1.0mm}%
\includegraphics[height=\h]{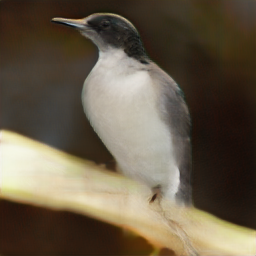}%
\end{minipage}
\caption{
Image generation based on condition space arithmetic of embedded textual descriptions.
\label{fig:regularities}
\vspace*{-1mm}
}
\end{figure}
}

\begin{document}

\title{Efficient Neural Architecture\\for Text-to-Image Synthesis
\thanks{This study was financed in part by the Coordenação de Aperfeiçoamento de Pessoal de Nivel Superior – Brasil (CAPES) – Finance Code 001.}}

\author{\IEEEauthorblockN{Douglas M. Souza, J\^onatas Wehrmann, Duncan D. Ruiz}
\IEEEauthorblockA{\textit{School of Technology} \\
\textit{Pontif\'icia Universidade Cat\'olica do Rio Grande do Sul}\\
Porto Alegre, Brazil \\
\texttt{\{douglas.souza90,jonatas.wehrmann\}@edu.pucrs.br, duncan.ruiz@pucrs.br}} 
% \and
% \IEEEauthorblockN{Jonatas Wehrmann}
% \IEEEauthorblockA{\textit{School of Technology} \\
% \textit{PUCRS}\\
% Porto Alegre, Brazil \\
% jonatas.wehrmann@edu.pucrs.br}
% \and
% \IEEEauthorblockN{Duncan D. Ruiz}
% \IEEEauthorblockA{\textit{School of Technology} \\
% \textit{PUCRS}\\
% Porto Alegre, Brazil \\
% duncan.ruiz@pucrs.br}
}

\maketitle

\begin{abstract}
Text-to-image synthesis is the task of generating images from text descriptions. Image generation, by itself, is a challenging task. When we combine image generation and text, we bring complexity to a new level: we need to combine data from two different modalities. Most of recent works in text-to-image synthesis follow a similar approach when it comes to neural architectures. Due to aforementioned difficulties, plus the inherent difficulty of training GANs at high resolutions, most methods have adopted a multi-stage training strategy. In this paper we shift the architectural paradigm currently used in text-to-image methods and show that an effective neural architecture can achieve state-of-the-art performance using a single stage training with a single generator and a single discriminator. We do so by applying deep residual networks along with a novel sentence interpolation strategy that enables learning a smooth conditional space. Finally, our work points a new direction for text-to-image research, which has not experimented with novel neural architectures recently.
\end{abstract}

\begin{IEEEkeywords}
text-to-image synthesis, generative models, multimodal learning.
\end{IEEEkeywords}

\section{Introduction}
Text-to-image synthesis is the task of generating images from text descriptions. Image generation, by itself, is a challenging task. When we combine image generation and text, we bring complexity to a new level: we need to combine data from two different modalities. In the most common setting, text-to-image methods are based on generative models that learn a text-conditioned distribution over images. Given a text description and some random variable, the algorithm produces a random image (controlled by the random variable) that correlates with the information present in the text. Text-to-image synthesis is a very recent research area and it has the potential to aid several real-world applications, from automated content generation to assisted drawing.

Most recent advances in text-to-image generation are driven by Generative Adversarial Networks (GANs) \cite{goodfellow2014generative}. GANs brought a leap of improvement in learning generative models over complex data distributions such as images. GANs have been successful in several tasks, such as image-to-image translation \cite{isola_2017_image,choi2017stargan,wang2017high,zhu_2017_unpaired}, image inpainting \cite{pathak_2016_context}, image editing \cite{zhu_2016_generative}, and image super resolution \cite{ledig_2016_photo}. In the context of text-to-image synthesis, a conditional GAN \cite{mirza2014conditional} is conditioned to a vector representation of the text description. In order to encode text to a vector representation, most methods rely on another algorithm, such as a Recurrent Neural Network (RNN) \cite{hochreiter1997long,cho2014learning}. There are two main levels in which a text description can be encoded to be used by GANs: sentence level and word level. At the sentence level, the entire description is encoded as global sentence vector. At the word level, on the other hand, there is a vector representation for each individual word in the description.

\figteaser
% Since this representation belongs to a continuous vector space, it makes text-to-image synthesis specifically different from traditional conditional GANs, which are usually conditioned to a discrete set of labels (such as class labels). 

% The most successful strategy was adding word-level features through Attention Mechanisms \cite{?} to help capture details from the text descriptions.

First approaches to text-to-image synthesis \cite{reed2016generative,reed2016generative,zhang2017stackgan,zhang2017stackgan++,zhang2018photographic} have simply extended GANs to be conditioned to sentence vectors. Naturally, results were not optimal. Most recent methods \cite{xu2018attngan,qiao2019mirrorgan,yin2019semantics,zhu2019dm} have proposed different strategies to circumvent the complex relationship between image and text. Most of those works, however, follow a similar pattern when it comes to neural architectures. Due to previously mentioned difficulties, plus the inherent difficulty of training GANs at high resolutions, most recent works have adopted a multi-stage training strategy. In a multi-stage setting, training is performed first at low resolutions (\ie~$32\times32$ and $64\times64$ pixels) and then refined to higher resolutions ($128\times128$ and $256\times256$ pixels). Usually, multi-stage training is implemented using several generators and several discriminators, which makes training complex and slow. This architectural choice has been followed by most previous work, which have been adding small improvements, such as word-level features through Attention Mechanisms~\cite{xu2018attngan}, Memory Networks~\cite{zhu2019dm}, Siamese Networks~\cite{qiao2019mirrorgan} and a Mirror strategy~\cite{qiao2019mirrorgan}.

In this paper, we shift the architectural paradigm currently used in text-to-image methods and show that an effective neural architecture can achieve state-of-the-art performance using a single stage training directly at the target resolution. By doing so, we not only introduce a simpler method for text-to-image synthesis but also point a new direction in text-to-image research, which has not experimented with novel neural architecture recently.

Specifically, we introduce an adversarial training-based architecture that leverages full capacity of modern deep convolutional networks, alongside to an improved sentence embedding approach for generating photorealistic text-conditioned images. Both discriminator and generator networks draw inspiration from~\cite{brock2018large}, though we provide important improvements on that architecture, allowing for the use of sentence embeddings rather than class labels as conditioning vectors. Results show that our models single-handedly outperform multi-stage state-of-the-art methods without heavy hyper-parameter optimization in two widely used benchmarks, namely CUB~\cite{WahCUB_200_2011} and Oxford-102~\cite{nilsback2008automated} datasets, in terms of both Inception Score~\cite{salimans2016improved} and Fr\'echet Inception Distance~\cite{heusel2017gans}. Fig. \ref{fig:teaser} shows samples generated by our method. Moreover, we provide an extensive set of experiments, in which we explore key components and abilities of our models.

Formally, in this work we make the following contributions:

\begin{itemize}
    \item We introduce a novel sentence interpolation strategy that allows the generator to learn a smooth conditional space, and also work as a data augmentation procedure.
    \item We show how the use of a modern residual neural architecture enables single-stage training at the target image size, and generates state-of-the-art text-to-image models.
    \item We perform an extensive analysis of the properties of text-to-image models, both in quantitative and qualitative fashion.
    \item We demonstrate that our models enable image editing using natural language via arithmetic operations in the conditional space, being able to modify aspects of the image while keeping its overall structure. 
\end{itemize}

\section{Related Work}

In this section we discuss the most important topics related to our work, namely, Generative Adversarial Networks (GANs), and specific methods designed to perform text-to-image synthesis.

\subsection{Generative Adversarial Networks}
Generative Adversarial Networks (GANs) \cite{goodfellow2014generative} is a class of generative method that learns generative models via an adversarial training procedure. In its traditional form, GANs are composed of two differentiable functions (\eg, neural networks), namely a Generator $G$ and a Discriminator $D$. $D$ is trained to correctly classify real and generated images while $G$ is trained simultaneously to make $D$ mistakenly classify generated images as real. 

Since its debut in 2014, there has been remarkable advances in GAN research. Important methods were proposed to address training stability and quality of results \cite{arjovsky2017wasserstein,mao2017least,gulrajani2017improved,karras2017progressive,miyato2018spectral,zhang2018self,mescheder2018training,brock2018large} and several tasks were improved by adversarial training, such as image-to-image translation \cite{isola_2017_image,choi2017stargan,wang2017high,zhu_2017_unpaired}, image inpainting \cite{pathak_2016_context}, image editing \cite{zhu_2016_generative}, and image super resolution \cite{ledig_2016_photo}.

\subsection{GANs for Text-to-image Synthesis}

Conditioning image generation of GANs on text descriptions was first proposed by Reed \etal \cite{reed2016generative} where the task was defined as two subtasks: encoding text descriptions to a vector representation and using this representation as a condition to train a Conditional GAN~\cite{mirza2014conditional}. In \cite{reed2016learning}, Reed \etal extends text-to-image to support location in which elements should be drawn. 

Zhang \etal \cite{zhang2017stackgan} proposed StackGAN, which introduced important concepts that are still used by recent works. StackGAN used stacked GANs to train text-to-image models in a two stage fashion: the first stage generates low resolution $64 \times 64$ pixel images then a second stage refines to a higher resolution of $256 \times 256$ pixels. StackGAN also introduced the Conditioning Augmentation (CA) module. CA maps text embeddings to a smooth known distribution that makes it easier to learn the text-to-image generator. Stackgan++ \cite{zhang2017stackgan++} extends StackGAN by adding multiples generators and discriminators, a pair is used at each of the following stages: $64\times64$, $128\times128$ and $256\times256$ pixels. HDGAN \cite{zhang2018photographic} follows similar multi-stage strategy but applies a patchwise adversarial loss. 

Since previous works had only used the global sentence embedding as condition to train text-to-image models, AttnGAN \cite{xu2018attngan} introduced Attention \cite{xu2015show} modules to add word-level cues so that generated images are closely related to the description. MirrorGAN \cite{qiao2019mirrorgan} learns text-to-image models by redescription, \ie~regenerating a text description for a generated image.
DM-GAN \cite{zhu2019dm} uses a dynamic memory module to select important aspects of first-stage images and refine to higher resolutions that are closely related to text descriptions. Finally, SD-GAN \cite{yin2019semantics} proposes a siamese multi-stage networks structure that is intended to make generated images consistent across a variety of descriptions. 

Our proposed method departs from most of the previously established strategies. We dramatically simplify the text-to-image framework. First, we shift the architectural paradigm from a multi-stage architecture to a single stage modern deep residual network, which makes training simpler and faster.
Second, we introduce a sentence interpolation strategy that allows the generator to learn a smooth conditional space, which not only improve results but also allow us to perform image editing by performing arithmetic operations in conditional space. Finally, we demonstrate that our method outperforms previous works that are also conditioned only on the sentence vector.

\section{Method}
In this section we present in details our proposed approach. Text-to-image synthesis methods have followed a similar design pattern regarding neural architectures: they make use of multi-stage training using several networks. This choice, however, increases training complexity and computational costs required to train such models. Our approach departs from this design altogether. We present evidence that the use of an adequate neural architecture plus a simple sentence interpolation strategy can produce state-of-the-art results. In addition, our method peforms a single-stage training with a single generator and a single discriminator. Next, we detail every component of our proposed method: the text encoder, the sentence interpolation strategy and the neural architecture. 

\subsection{Text Encoder}

We encode text descriptions into a vector representations by using a pre-trained Deep Attentional Multimodal Similarity Model (DAMSM)~\cite{xu2018attngan}. The DAMSM module, similarly to~\cite{wehrmann2020aaai,faghri2018vse,wehrmann2019iccv,lee2018stacked,wehrmann2018cvpr}, learns image and text encoding functions, namely $\varphi(\image)$ and $\phi(\sentence)$, that map images $\image$ and textual descriptions $\sentence$ into the same semantic multimodal space. Such a space is trained so that correlated image-caption pairs lie close to each other, while non-correlated pairs must present larger distance than the correlated ones. By optimizing that space, the learned text representation is forced to closely resemble the content from images, and therefore can be as a condition vector $\sentemb \in \mathbb{R}^{256}$ in our architecture. 

Original image captions $\sentence$ are tokenized, and each token is represented by a specific vector $\mathbb{R}^{300}$. Those vectors feed a Bidirectional GRU network, which provides per-token hidden representations, as well as a final global vector. Hidden representations are used for learning fine-grained correlations with the spatial information of the images, while the global vector contains holistic high-level information of the caption. In this study, we use the global vector alone as textual condition vector $\sentemb$, hence $\phi(S)=\sentemb$.

% Falar da atenção? Global-local? 
% Colocar o loss? 
% ?

% \begin{equation}
%     \varphi \left(\text{\textit{``This is a red bird"}} \right) - \varphi \left(\text{\textit{``It is red"}} \right)
%     + \varphi \left(\text{\textit{``It is blue"}} \right)
% \end{equation}

\subsection{Sentence Interpolation}

In this section we detail a novel strategy for improving the smoothness of the conditional space, which we hereby call Sentence Interpolation (\si). This technique consists in using all the available captions for computing the general sentence embedding regarding an image during training. By doing so, we make the textual representation vector to be continuous in the projected space, rather than being discrete points in the manifold, as a traditional approach would generate. 

Formally, let $\image_i$ be the $i^{th}$ image from the training dataset, and $S_{ij} = \{\sentemb_1, \sentemb_2, ..., \sentemb_n \}$ be the set of $n$ correlated sentence embeddings that describe that particular image. We sample an $n$-sized vector of weights $\m \sim \mathcal{U}(0, 1)$, and further normalize it with a softmax function. Those normalized values are used to weight each one of the sentence vectors, so their sum consists in an interpolated representation of the original sentences. Therefore, the vector $\dot{\sentemb}$ that represents the interpolated textual embedding of a given image is calculated as follows:

% \begin{equation}
%     \m = \Bigg (\frac{e^{\m}}{\sum_{i=1}^{n}e^{\m_{i}}} \Bigg ) 
% \end{equation}

\begin{equation}
    \dot{\sentemb} = 
        \sum_{j=1}^{n} \Bigg [ 
            S_{j} \times \bigg (
                \frac{e^{\m}}{\sum_{k=1}^{n}e^{\m_{k}}} \bigg 
            )_j 
        \Bigg ]
\end{equation}

Such an approach makes a limited set of sentences to be represented by countless continuous points during the training process. The main implications of this technique are two-fold: (i) it makes the sentence embedding space to be more smooth; (ii) and also works as a data augmentation strategy, given that the same textual descriptions can assume different forms depending on the sampling of $\m$. In comparison to the Conditioning Augmentation (CA) module introduced by StackGAN \cite{zhang2017stackgan}, the sentence interpolation has the advantage of being deterministic. This is due to the fact that it is not used during the test phase. CA, on the other hand, introduces randomness when encoding sentence vectors during training and testing.

\subsection{Architecture}
We follow the steps of Brock \etal \cite{brock2018large}, which introduced the state-of-the-art architecture for GANs, namely BigGAN-Deep. This architecture is based on residual blocks with bottleneck structure of He \etal \cite{he2016deep}, which makes deeper networks more computationally efficient and easier to train. Also, like SAGAN \cite{zhang2018self}, BigGAN-Deep applies Spectral Normalization \cite{miyato2018spectral} and Non-local Blocks \cite{wang2018non} in both generator and discriminator. Finally, BigGAN-Deep introduces conditioning information in the generator using Conditional Batch Normalization \cite{dumoulin2016learned} and in the discriminator using the projection approach of Miyato \etal \cite{miyato2018cgans}.

BigGAN-Deep presented a new state-of-the-art result in the ImageNet \cite{imagenet_cvpr09} dataset in the supervised setting. Therefore, it was designed to be conditioned on class labels. Since in this architecture class labels are represented by dense embeddings, we extended it to handle the sentence vector. Specifically, we replaced the trainable class embeddings by the fixed sentence vectors $\sentemb$. In the discriminator, sentence vectors are linearly projected to be used in the projection conditioning. In the generator, sentence vectors are concatenated with the noise vector $\bmz$ and then linearly projected to form BatchNorm gains and biases, gains are one-centered while biases are zero-centered. By using the fixed sentence vectors, the generator and discriminator are forced to adapt to the conditional space learned by the DAMSM encoder, which yields interesting properties, such as the generator's ability to handle arithmetic operations in conditional space, which is presented in Section \ref{sec:cond_arithmetic}.

The BigGAN-Deep architecture was originally designed to be used in large scale training. Large scale training is done by using a big batch size (\eg~2048) and training the models across several devices. In order to apply this architecture in a small scale, we need to make additional adaptations. First, we switch relu activation to leaky relu. This helps avoiding sparse gradients, which is helpful due to the second adaptation. Second, we reduce the number of parameters of both networks. We reduce the number of parameters in the generator and discriminator by reducing the channel multiplier $ch$ to 96 instead of 128 in default BigGAN-Deep architecture. This reduction represents 43\% less parameters in the discriminator and 36\% less parameters in the generator. Finally, training is performed directly at the target resolution of $256\times256$ pixels. As far as we know, no previous text-to-image method was able to train directly at this resolution without relying on multiples generators and discriminators.

\subsection{Objective Function}
We adopt the so-called hinge adversarial loss. The hinge loss works similar to WGAN loss~\cite{arjovsky2017wasserstein} but is more stable thanks to the margins introduced in the discriminator loss function. For the discriminator, the hinge loss is given by:

\begin{align}
V_D(\hat{G}, D) &= \E_{\bmx,\sentemb\sim q_{\rm data}}\left[{\rm min}\left(0, -1+D(\bmx, \sentemb)\right)\right] + \\
	 & \E_{\bmz\sim p_{\bmz},\sentemb\sim q_{\rm data}} \left[{\rm min}\left(0, -1-D\left(\hat{G}(\bmz, \sentemb),\sentemb\right)\right)\right], \notag
 \label{eq:hinge}
\end{align}
where $\bmx$ and $\sentemb$ are real images and their corresponding sentence vectors, respectively. $\hat{G}(\bmz, \sentemb)$ is a fake image from the generator for a given random vector $\bmz$ and a sentence vector $\sentemb$, respectively. Note that the hat in $G$ means that, in this case, the generator's weights are not being updated.
 
Similarly, the loss function for the generator is given by:

\begin{align}
V_G(G, \hat{D}) &= -\E_{\bmz\sim p_{\bmz},\sentemb\sim q_{\rm data}}\left[\hat{D}\left(G(\bmz,\sentemb), \sentemb\right)\right],  
\end{align}

in this case, the hat in $D$ means the discriminator's weights are not being updated.

\section{Experiments}
In this section we present our experimental setup. We conduct extensive experiments in the most used datasets for text-to-image generation. We also present an extensive quantitative and qualitative analysis of our findings.

\subsection{Datasets}
We have used two widely used datasets for training and evaluating our models, as follows.

\textbf{Caltech-UCSD Birds (CUB)~\normalfont{\cite{WahCUB_200_2011}}:} The CUB dataset is composed of 11,788 images of birds distributed among 200 class categories. The dataset is split in 8,855 images of 150 categories for training and 2,933 images of 50 categories for testing. Each image contains 10 text descriptions.

\textbf{Oxford-102~\normalfont{\cite{nilsback2008automated}}:} The Oxford-102 dataset is composed of 8189 images of flowers of 102 categories. The dataset is split in 7034 images for training and 1154 images for testing. Each image contains 10 text descriptions.

\figarithmetic

\subsection{Evaluation}
In order to evaluate our method, we employ the two most widely used metrics to evaluate generative models: the Inception Score (IS) and the Fr\'echet Inception Distance (FID). The IS uses a pretrained Inception Network \cite{szegedy2015going} to compute class probabilities over generated samples. IS is both a measure of \textit{objectness} and variety, therefore, the higher the score the better. In order to compute IS, and also be able to compare results, we use the same Inception Networks used to evaluate previous work. The networks are provided by StackGAN~\cite{zhang2017stackgan} and are finetuned for the CUB and Oxford-102 datasets.

\begin{table*}[!t]
\caption{Quantitative comparison of text-to-image methods.}
\begin{center}
\begin{tabular}{lccccccc}
\hline
\multirow{2}{*}{Method} & \multicolumn{2}{c}{\# Networks} & \multirow{2}{*}{Multi-Stage} &\multicolumn{2}{c}{\IS} & \multicolumn{2}{c}{\FID}\\
% \cline{2-3}
             & Discriminators & Generators & &  CUB   & Oxford-102 & CUB & Oxford-102\\
\hline\hline
GAN-INT-CLS~\cite{reed2016generative} & 1 & 1 & No &  2.88 $\pm$ 0.04  & 2.66 $\pm$ 0.03 & - & - \\
GAWWN~\cite{reed2016learning}         & 1 & 1 & No &  3.60 $\pm$ 0.07  & - & - & - \\
StackGAN~\cite{zhang2017stackgan}     & 2 & 2 & Yes &  3.70 $\pm$ 0.04  & 3.20 $\pm$ 0.01 & 55.28 & 51.89 \\
StackGAN++~\cite{zhang2017stackgan++} & 3 & 3 & Yes &  4.04 $\pm$ 0.05  & 3.26 $\pm$ 0.01 & 15.30 & 48.68\\
TAC-GAN~\cite{dash2017tac}            & 1 & 1 & No  &  - & 3.45 $\pm$ 0.05 & - & - \\
HDGAN~\cite{zhang2018photographic}    & 3 & 3 & Yes &  4.15 $\pm$ 0.05  &  3.45 $\pm$ 0.07 & - & - \\
\hline
Ours           & 1 & 1 & No &  \textbf{4.23 $\pm$ 0.05} &  \textbf{3.71$\pm$ 0.06} & \textbf{11.17} & \textbf{16.47}\\
\hline
\label{tab:results}
\end{tabular}
\end{center}
\end{table*}

A downside of the IS is that it does not consider the statistics present in the real data. A generative model that generates a few high quality examples for each class would have a very high IS score, despite its variety being low. To circumvent this issue, Heusel \etal \cite{heusel2017gans} introduced the Fr\'echet Inception Distance (FID). FID considers the statistics present in the training data, so it possibile to evaluate if the generative model learned a distribution that have similar statistics. Specifically, FID uses an Inception Network to compute activation features of both training set images and generated images. The Fr\'echet Distance is then computed over the features of real and fake images. FID gives a measure of how close the statistics of generated images are from those in the training set, hence, the lower the score the better.

\begin{figure}[!tpb]%
\centering
{\includegraphics[width=0.99\columnwidth]{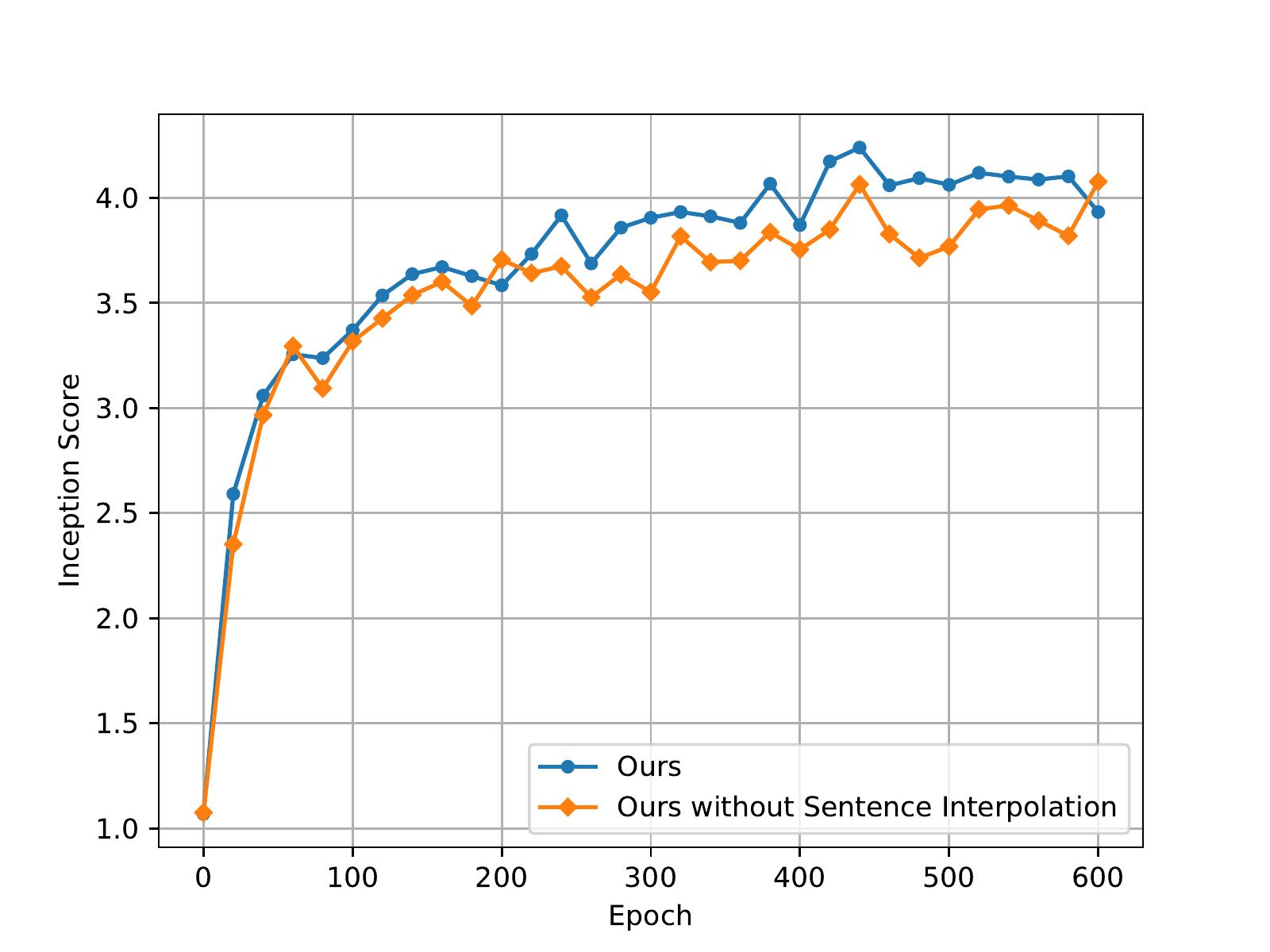}}
\caption{Inception Score during training epochs for our model with and without Sentence Interpolation in the CUB dataset.}
\label{fig:is_comparison}
\end{figure}

\subsection{Implementation Details}

We use Adam optimizer \cite{kingma2014adam} with a learning rate of $4 \times 10^{-4}$ for $D$ and $10^{-4}$ for $G$. We set $\beta_{1}=0$ and $\beta_{2}=0.999$ for both $G$ and $D$. We train one $D$ step per $G$ step. We use synchronized implementation of BatchNorm, where statistics are aggregated across all devices. We keep an exponential moving average of the generator weights with a decay of $0.999$ for sampling. Since BatchNorm statistics are not computed for averaged weights of the generator, we employ the ``standing statistics" strategy of Brock \etal \cite{brock2018large}. In other words, we first run 100 forward passes through $G$ to update its BatchNorm statistics, making the generator invariant to batch sizes. Finally, we perform training using 3 GPUs with a batch size of 8 per GPU, making up for a batch of size 24. Most models take up to 3 days to train.

\section{Comparison to state-of-the-art methods}

In order to provide reassurance on the generative performance of our models, we compare their quantitative and qualitative results against current state-of-the-art methods~\cite{reed2016generative,zhang2017stackgan,zhang2017stackgan++,dash2017tac,zhang2018photographic}. Note that some of them have not reported FID results. Hence, we compare to the results publicly available.

\subsection{Quantitative Analysis}

Table~\ref{tab:results} depicts quantitative results, alongside to the number of discriminator and generator networks used in each work. It arguably shows that our approach is the preferred method, once it achieves top performance in all metrics while employing just a single discriminator and a single generator in the entire architecture. Notably, it outperforms all the baseline approaches by a margin across all datasets and metrics. 

The largest improvement provided by our approach is on Oxford-102 dataset. It provides a relative improvement of $\approx 7\%$ IS and $\approx 300\%$ FID when compared to the strongest baseline with public results available. Clearly our approach also leads to a significantly better results on CUB dataset, allowing for a $\approx 24\%$ FID reduction. 

\subsection{Qualitative Analysis}

Fig.~\ref{fig:compare_birds} depicts qualitative results of models trained on CUB dataset. In that Fig., we compare our model to the baseline ones. One can observe that our model brings improvement on several aspects regarding the generated images. For instance, our images look more photorealistic, present better semantic correspondence of the generated images to the provided description, and also generate more fine-grained details in both foreground and background elements.

Results shown in Fig.~\ref{fig:compare_flowers}~were generated using a model trained on Oxford-102 dataset. Once again, our model generates images with much richer detail level and photorealistic aesthetic. Such experiment supports our claims that our proposed single-stage architecture can be used for generating concepts across distinct datasets. It is worth noticing that despite Oxford-102 being a somewhat small dataset, our models were able to learn a proper distribution without suffering from mode collapse or additional training instabilities. 

\figbirds
\figflowers

\section{Condition Space Arithmetic}
\label{sec:cond_arithmetic}
In this section we explore the inherent capability of our approach to handle condition space arithmetic. This is a very interesting property and finds applications in many real world tasks, such as image manipulation via natural descriptions. This capability emerges from the fact that the employed sentence embedding vector $\sentemb$ concatenated to the $z$ vector lie in a smooth embedding space that present structural regularity. In that particular kind of space we can find semantic regularities regarding concepts learned by the model, i.e., they respect a semantic organization of concepts. We observed that the use of our novel sentence interpolation strategy during training is quite helpful to improve the learned condition space. It increases the model capacity of learning a smooth condition space, in which embedding regularities are more easily found. 

Fig.~\ref{fig:regularities} showcases examples regarding regularities found in our trained models. For generating those images we hold $z$ fixed, and embed captions into the multimodal space, which are used in simple vector operations, as follows. The uppermost example depicts an image generated by $G(z, \phi(``\text{\textit{This is a red bird}}"))$. We then subtract $\phi(``\text{\textit{It is red}}")$ from $\phi(``\text{\textit{This is a red bird}}")$, and generate a novel image (in the center). One can see that such operation completely removed the red color from the generated bird. Finally, we add $\phi(``\text{\textit{It is blue}}")$ to the resulting embedding, and use it to generate the rightmost image. That image shows the same bird, though with its color changed from red to blue, using only simple vector-level operations. 

Note that our models are able to edit images while preserving the main image structural content without even being explicitly trained to learn disentangled representations. Fig.~\ref{fig:regularities} also demonstrates that one can edit several aspects of the generated images, such as shape of the beak, and presence of colored crown.

\begin{figure}[!tpb]%
\centering
\subfloat[Sentence embeddings sampled without Sentence Interpolation.]{\includegraphics[width=0.5\columnwidth]{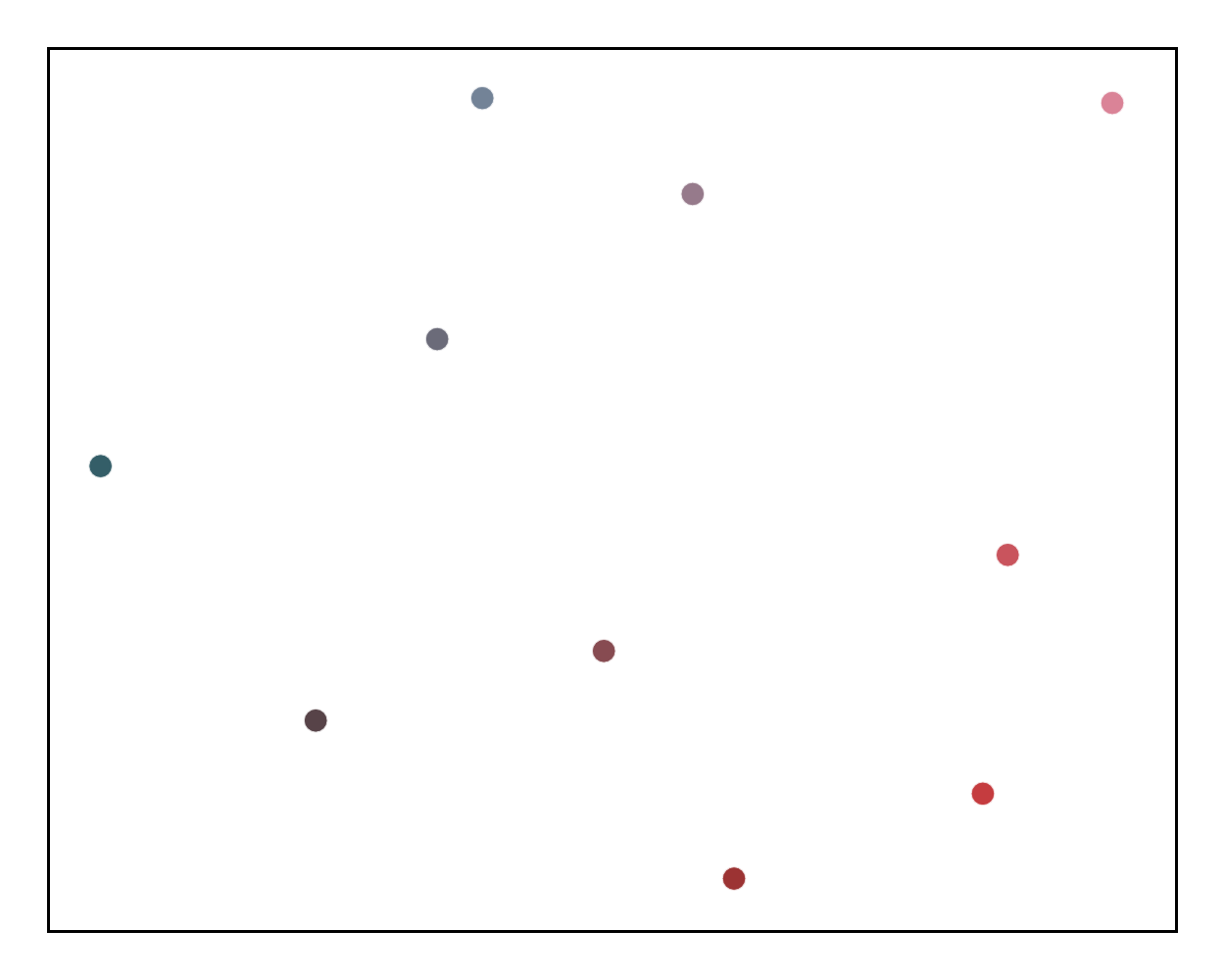}}\\
\subfloat[Sentence embeddings sampled with Sentence Interpolation.]{\includegraphics[width=0.5\columnwidth]{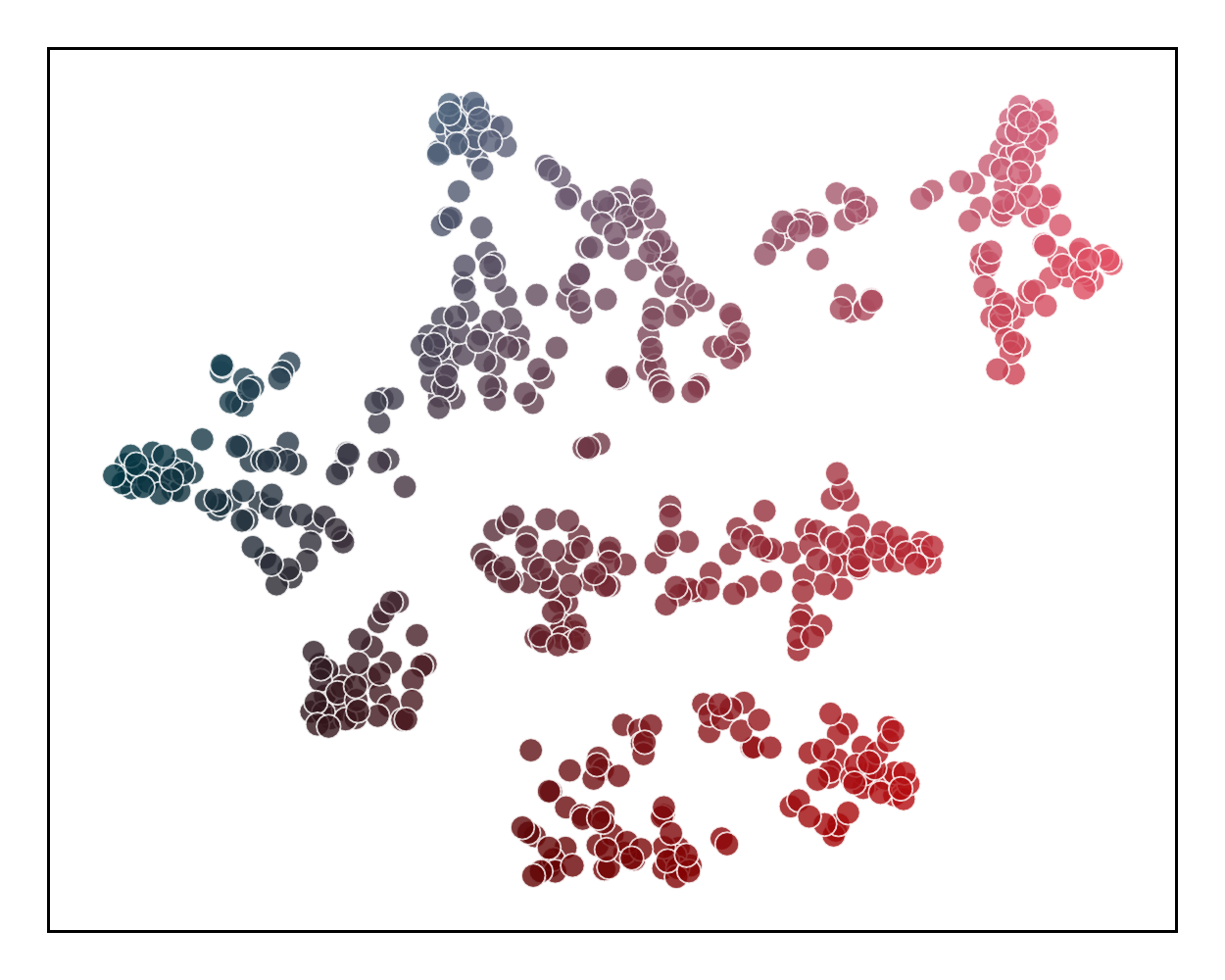}}
\caption{Manifold visualization of the sampled sentence embeddings during training. We visualize sentence embeddings by applying t-SNE~\cite{maaten2008visualizing} to project sentence embeddings from the original $\mathbb{R}^{256}$ space to a $\mathbb{R}^{2}$ space. We show 10 sentence embeddings of a randomly chosen image during the entire training (\ie, resulting in 600 embeddings). In (a) is shown the regular sampling of a random sentence. In (b) is shown the sampling using the Sentence Interpolation.}
\label{fig:manifold}
\end{figure}

\section{Impact of Sentence Interpolation}

One of the contributions of this paper regards the introduction of a novel Sentence Interpolation procedure. In order to understand its effects, we have trained two models: (i) a default complete model that performs Sentence Interpolation; and (ii) a model with the same overall architecture, though without applying any interpolation between sentences. Fig.~\ref{fig:is_comparison}~shows per-epoch Inception Score values computed during the entirety of the training process. It arguably proves the importance of the proposed technique. During the early stages of training, results are indeed quite similar, the difference being more relevant after the $100^{th}$ epoch. Notably, after the $400^{th}$ epoch, IS results with Sentence Interpolation were consistently higher than 4.00, while the model without it surpassed that mark only twice throughout the training. 

Effects of the \si~approach also can be seen in Fig.~\ref{fig:manifold}. In this analysis, we plot ten sentence embeddings of a randomly chosen image during the entire training (\ie, resulting in 600 embeddings). We plot the very same embeddings for the model trained with and without \si. We apply the t-SNE~\cite{maaten2008visualizing} technique on those embeddings so as to project $\mathbb{R}^{256}$ vectors onto a $\mathbb{R}^2$ space. Such a plot clearly shows that the proposed interpolation provides a much larger exploration of the sentence embedding manifold, allowing for sampling continuous points from that space. That sampling region is obviously constrained by the ten points regarding the image descriptions chosen. 
We intend to further extend this technique for future work, so as to allow sampling points from outside of those boundaries, without loosing semantic context. 
When training without it, one can only sample fixed discrete points, which poses a considerable constraint on the information carried on the condition vector. This analysis corroborates with our hypothesis that \si~works also as a data-augmentation scheme, providing better generation results for points present in a larger region of the manifold.

\begin{figure}[!tpb]%
\centering
{\includegraphics[width=0.9\columnwidth]{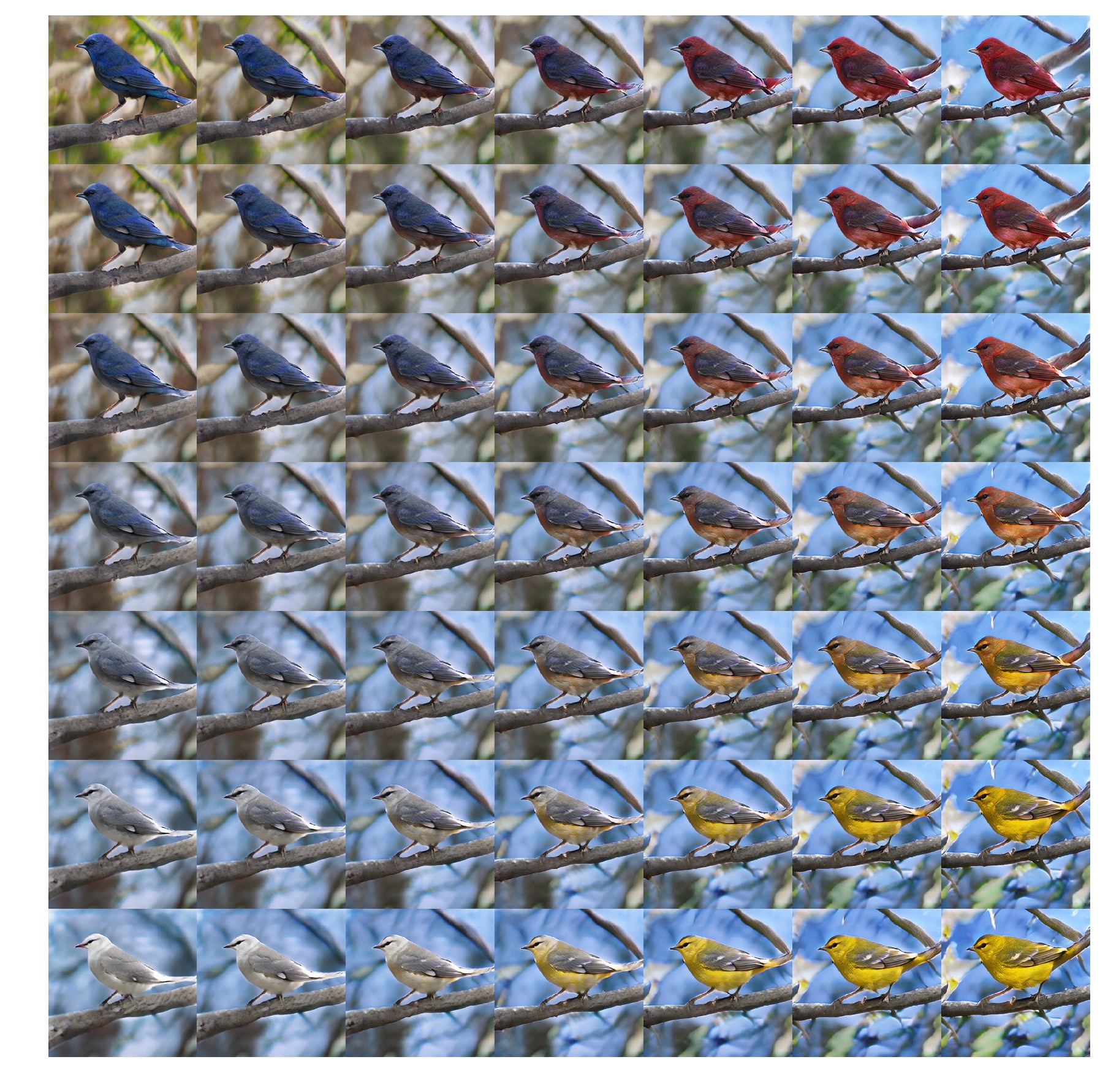}}
\caption{Image generation with sentence embeddings linearly interpolated across all directions. There are four original embeddings, each one used to generate an image (those from the four corners), while all the remaining ones were generated using interpolated description embeddings. The upper-left position depicts an image generated with the description ``It is a blue bird", the bottom-left image was generated with ``It is a white bird", the upper-right image with ``It is a red bird", and the bottom-right image with ``It is a yellow bird".}

\label{fig:captioniterp}
\end{figure}

\section{Conclusion and Future Work}

In this work, we propose a novel approach that shifts the architectural paradigm currently used in text-to-image methods. We show that an effective neural architecture can achieve state-of-the-art performance using a single stage training directly at the target resolution. By doing so, we not only introduce a simpler method for text-to-image synthesis but also point a new direction in text-to-image research.

In a future work we intend to explore different ways of introducing the sentence vector as condition in the discriminator. Since the projection conditioning introduced by Miyato \etal \cite{miyato2018cgans} was designed to work in conjunction with trainable class embeddings, we believe that there is space to be explored when the condition is a fixed sentence vector. Additionally, we would like to investigate the impact of adding components of recent works, such as Attention Modules~\cite{xu2018attngan}~and Memory networks \cite{zhu2019dm}.

\bibliographystyle{IEEEtran}
\bibliography{main}

\end{document}